# Efficient Multi-branch Segmentation Network for Situation Awareness in Autonomous Navigation

Guan-Cheng Zhou[a], Chen Cheng[b], Yan-zhou Chen[a]

[a]School of mathematics and statistics, Xi'an Jiaotong University Xi'an, 710000, China

[b]School of Naval Architecture and Ocean Engineering, Jiangsu University of Science and Technology Zhenjiang, 212013, China

**Abstract**

Real-time and high-precision situational awareness technology is critical for autonomous navigation of unmanned surface vehicles (USVs). In particular, robust and fast obstacle semantic segmentation methods are essential. However, distinguishing between the sea and the sky is challenging due to the differences between port and maritime environments. In this study, we built a dataset that captured perspectives from USVs and unmanned aerial vehicles in a maritime port environment and analysed the data features. Statistical analysis revealed a high correlation between the distribution of the sea and sky and row positional information. Based on this finding, a three-branch semantic segmentation network with a row position encoding module (RPEM) was proposed to improve the prediction accuracy between the sea and the sky. The proposed RPEM highlights the effect of row coordinates on feature extraction. Compared to the baseline, the three-branch network with RPEM significantly improved the ability to distinguish between the sea and the sky without significantly reducing the computational speed.

# 1. Introduction

Tourism and maritime industries are thriving again in the information age. With increasing labour costs, advanced automated vessels are becoming crucial in maritime operations. Unlike other marine equipment, unmanned surface vehicles (USVs) are low-cost, safe to operate, and highly flexible (Liu et al., 2016). Meanwhile, the computer vision field has significantly developed with the advancement of artificial intelligence, which helps USVs gain improved awareness of the surrounding environment.

With the advancement of deep learning technology, vision-based USV situation awareness technology has received increasing attention. Deep convolution has improved the capabilities of traditional neural networks in object detection. Zou et al. (2020) proposed a novel three-step approach for water-shore-line detection in boat-borne vision images using image sequence information. Lee et al. (2018) proposed neural network-based object detection algorithms to identify ships in images and videos captured at sea. Using deep learning methods, Zhao et al. (2019) and Yin et al. (2022) proposed a novel method based on 1-stage object detection and proved that this method outperformed traditional methods based on 2-stage object detection, such as Region with Convolutional Neural Network (R-CNN) and Fast R-CNN (Ren et al., 2015), regarding efficiency and accuracy.

Unlike object-detection convolutional methods, segmentation-based convolutional methods perform better in detecting general obstacles. Cane and Ferryman (2018) evaluated segmentation networks in a filtered segmentation ADE20k dataset (Zhou et al., 2017). Moreover, Bovcon et al. (2019) built a marine semantic segmentation training dataset tailored for obstacle detection methods in small-sized coastal USVs. Both studies indicated accuracy limitations in water segmentation and the misclassification of small obstacles. Consequently, designing a segmentation architecture specifically for maritime environments was recommended. Steccanella et al. (2020) improved the traditional U-shape segmentation network (U-Net, Ronneberger et al., 2015) architecture with depth-wise convolutional layers.

However, the perception range of USVs is limited by the installation height of the sensors, which affects the safety and efficiency of daily marine operations. Thus, unmanned aerial vehicles (UAVs) have been integrated with the perception systems of USVs to solve these limitations. Moreover, their perception ranges have been improved (Wang et al., 2023). Currently, these integrated systems have focused on increasing research attention in the robotics and automation fields. In particular, a single-sensor input is insufficient to perceive the surrounding environment of the USV better. Thus, USV–UAV collaborative systems have gained increasing attention. The system proposed by Xu et al. (2019) achieved UAV landing on a moving target on a USV by recognising cooperative target points and performing pose estimations. To fully leverage the sensing advantages of UAVs in the control of USVs, Wang et al. (2023) proposed a cooperative USV–UAV system based on visual navigation and reinforcement learning-based control for marine search and rescue.

Despite the similarity in the visual information between the perspectives of USVs and UAVs, differences remain. The effect of sea fog on visibility becomes more severe as altitude increases; thus, distinguishing between the sea and sky and detecting small obstacles in images is challenging. Maritime ports differ from traditional wharf environments. In particular, vessels in ports, such as cargo ships and construction vessels, are typically larger. Meanwhile, construction vessels contain large towers, highly similar to cranes. These findings indicate that USV and UAV maritime port environments require a more efficient architecture.

We built a dataset for training USV and UAV semantic segmentation in a maritime port environment. In maritime environments, most views captured by USVs and UAVs comprise the ocean, land, and sky. In practical applications of deep learning methods, distinguishing between the sea and sky is challenging. By counting the row index and pixel labels (Figure 1), the number of pixels labelled as the sea in each row is positively correlated with the row coordinates, whereas the number of pixels labelled as the sky is negatively correlated. Hence, from a statistical perspective, pixels with higher row coordinates are more likely to correspond to the sea. In contrast, those with lower-row coordinates are more likely to correspond to the sky. The experimental data can only be used for academic purposes, and researchers can contact us to obtain the experimental data by email.

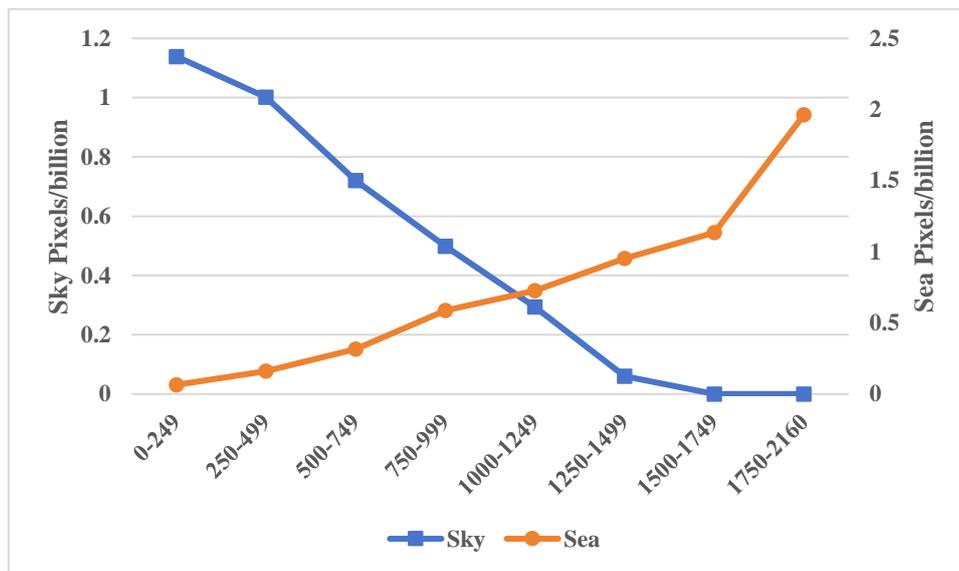

Figure 1. Count result. The X-axis represents the range of row coordinates, and the Y-axis represents the number of pixels belonging to that category.

These findings highlight the relevancy of row position information in the feature extraction process. Inspired by positional encoding in transformers (Vaswani et al., 2017), absolute position encoding can be deployed in a convolution network to emphasise the effect of row coordinates on the features. Based on this concept, we propose a novel row positional encoding module (RPEM). Specifically, position encoding was constructed based on row coordinates and incorporated into features with weights based on the input. Xu et al. (2023) proposed a three-branch semantic segmentation network inspired by proportional-integral-derivative controllers (PIDNet) to improve boundary accuracy in semantic segmentation. By adding a boundary branch, the network achieved a higher edge detection accuracy while improving the detection rate of small objects, aligning well with the requirements of maritime port segmentation. Thus, we followed PIDNet and constructed our network as a three-branch semantic segmentation network to improve the accuracy of waterlines. The source code can be accessed via: https://github.com/GuanchengZhou/PIDNet-with-RPEM.

The main contributions of this study are as follows:

• A three-branch semantic segmentation network with RPEM was proposed to improve the prediction accuracy of sea, sky, and small obstacles.

• A novel RPEM was proposed to emphasise the effect of row coordinates on feature extraction.

• A dataset was built to train both USV and UAV semantic segmentation in the maritime port environment, containing the data from the USV and UAV perspectives.

The organisation of this paper is as follows. In Section 2, studies related to segmentation technology in USV autonomous navigation are introduced. Section 3 introduces the details of the proposed method. Subsequently, Section 4 demonstrates the experiment setup and results. Finally, the conclusions are presented in Section 5.

## 2. Related Work

Advancements in camera technology have enabled the development of powerful and affordable sensing devices. Although autonomous marine robotics is a relatively new field, numerous image-processing methods have been designed to detect obstacles and waterlines.

Semantic segmentation classifies pixels into different categories. This technology can be employed for tasks related to autonomous navigation, such as sea–sky segmentation and obstacle detection. Segmentation-based methods perform better than detection-based methods for detecting waterlines and general obstacles. However, semantic segmentation requires more supervised data compared with object detection tasks. A marine semantic segmentation training dataset (MaSTr1325 ,

Bovcon et al., 2019) was captured in a maritime environment and annotated for each pixel. Žust and Kristan (2022) incorporated images from inland environments into MaSTr1325 and proposed MaSTr1478. Žust et al. (2023) proposed a Lakes Rivers and Seas dataset (LaRS), whose data contained images captured from lakes, rivers, and seas. However, all these datasets only collected data from the perspective of USVs. In contrast, although the perspective from UAVs shares similarities with that of USVs, distinguishing between the sea and the sky with UAVs is challenging. Therefore, we built a segmentation dataset in a maritime port environment to train both USV and UAV models.

Traditional single-branch semantic segmentation networks have been widely applied to maritime environments. Kim et al. (2019) proposed Skip-Enet, which applied skip-connections and whitening layers to the efficient neural network for semantic segmentation(E-Net, Paszke et al., 2016) to improve the capability of predicting obstacles. Steccanella et al. (2020) improved the traditional U-Net (Ronneberger et al., 2015) architecture with depth-wise convolutional layers. Bovcon and Kristan (2021) proposed Water Segmentation and Refinement Maritime Obstacle Detection Network (WaSR), which contains deep encoder–decoder architecture, and added IMU data to assist the segmentation task. However, contextual dependencies can be captured only through a large receptive field, implying a higher computational cost.

Unlike a single-branch segmentation network, a two-branch network utilises a context branch to fulfil the contextual dependency for semantic segmentation. Yu et al. (2018) proposed Bilateral Segmentation Network for Real-time Semantic Segmentation (BiSeNet), which contains a feature fusion module to mix the information from two branches. The model was then further developed, and BiSeNetV2 (Yu et al., 2021) was proposed; the modified model contained a guided aggregation layer to enhance mutual connections. Deep Dual-resolution Networks (DDRNet, Hong et al., 2021) included a bilateral connection to enhance the information exchange between the detail and context branches. However, large object blurring causes unclear boundaries for small objects in the traditional two-branch segmentation network; thus, Xu et al. (2023) proposed PIDNet. This network contains a boundary branch that uses boundary information to assist the fusion of the detailed and context branches. Here, to better distinguish between the sea and the sky, we used PIDNet and proposed a three-branch semantic segmentation network with RPEM.

## 3. Methods

We propose a three-branch semantic segmentation network with an RPEM to improve the prediction accuracy for the sea, sky, and small obstacles. The RPEM emphasises the effect of row coordinates on feature extraction. A multitask training method was proposed to train the multi-branch network.

## 3.1 Three-Branch Semantic Segmentation Network with RPEM

Following PIDNet (Xu et al., 2023), our network was developed as a three-branch segmentation network (Figure 2) comprising stem, backbone, and head networks. The stem network rapidly downsamples the image and reduces the computational cost of subsequent processing. The backbone network was used to extract the features of the inputs. The head network maps the feature space to the output space. The stem and backbone networks were the main components used to extract features from the inputs; therefore, we only added RPEM to the stem and backbone networks. In addition, the prediction head considered the traditional fully convolutional network (FCN) (Long et al., 2015) structure.

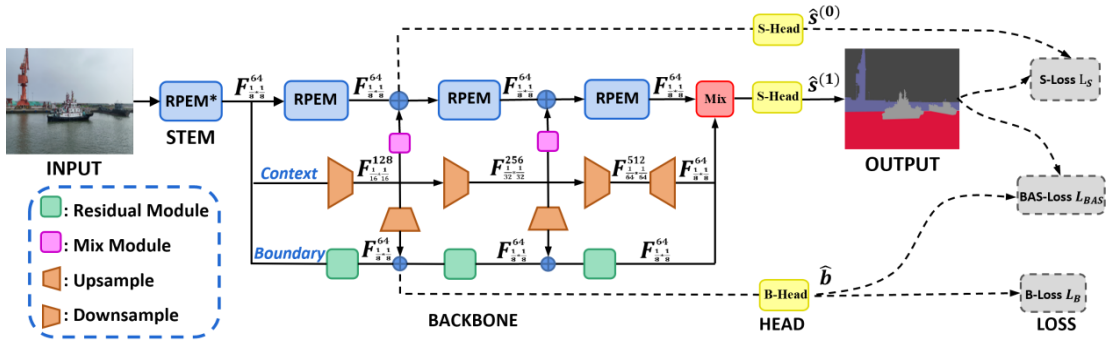

Figure 2: Structure of the proposed network. The branches in these three rows are the detail, context, and boundary. The symbol $F$ represents the tensor of the feature. The RPEM* downsamples the input while maintaining the origin size of the input.

The backbone network contains three branches dedicated to parsing detailed, contextual, and boundary information. Contrary to traditional two-branch semantic segmentation networks (Hong et al., 2021; Yu et al., 2018; Yu et al., 2021), the proposed network comprised an added boundary branch to predict boundary information. Finally, the three branches of the network are the detail, context, and boundary branches, which focus on local, surrounding, and boundary information, respectively. Notably, the context branch assists the detail and boundary branches in feature extraction. During prediction, the context branch continuously provides context auxiliary features to the detail and boundary branches. Moreover, position information is highly correlated with detailed information; hence, we only added the RPEM to the backbone and detail branches.

Because of the three-branch structure of the backbone network, three prediction heads exist. Prediction heads can be divided into two categories based on the task: semantic heads (SHead) and boundary heads (B-Head). S-Head is designed to predict the class of each pixel based on the input features from the midline detail branch and the backbone and generate the semantic segmentation output ($\hat{s}^{(0)}$ and $\hat{s}^{(1)}$). On the other hand, B-Head is designed to predict whether each pixel is an edge by processing the input features and ultimately outputs a probability map ($\hat{b}$). Both S and B-heads use the traditional FCN (Long et al., 2015) structure to balance computational speed and accuracy.

**3.2 Row Positional Encoding Module (RPEM)**

The RPEM was proposed to help the convolution network learn more useful features and row positional information. It combines the input $F_{in}$ with row position encoding (RPE) to obtain an output $F_{out}$, whose structure is shown in Figure 3. The underlying concept of this module was based on attention mechanisms (Vaswani et al., 2017). For the module input $F_{in}$, the output of the sigmoid function can be represented as follows:

$$\sigma = Sigmoid(f_\sigma(f(F_{in}))) \tag{1}$$

where $f_\sigma$ is a traditional convolution module function, $f$ is a function of n residual blocks, and $\sigma$ indicates the importance of the positional information based on $F_{in}$. If $\sigma$ is high, the row information is more significant. Thus, the output of this module can be written as follows:

$$F_{out} = \sigma f_{pos}(f(F_{in}))RPE + (1-\sigma)f(F_{in}) \tag{2}$$

where $f_{pos}$ is a conventional convolutional module function that differs from $f_\sigma$.

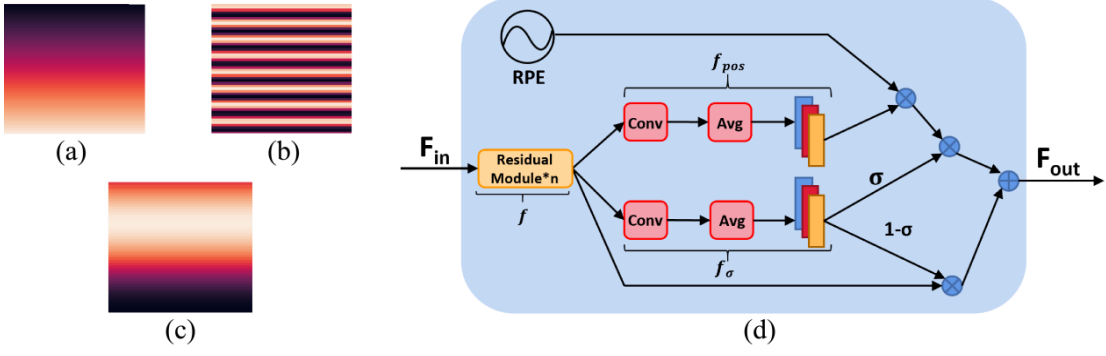

Figure 3. Visualisation of positional encoding of the first dimension and PEM. (a): linear positional encoding. (b): sine positional encoding. (c): sine positional encoding with normalise. (d): structure of RPEM, where RPE represents row position encoding.

In particular, we used similar types of positional encoding for the transformer and linear positional, including sine positional encoding (Vaswani et al., 2017), as shown in Figure 3. The linear positional encoding of a $d_{model}$-dimensional feature is calculated as follows:

$$RPE(i, j, d) = \frac{i}{m} \qquad (3)$$

The sine positional encoding of a $d_{model}$-dimensional feature can be calculated as

$$RPE(i, j, 2d) = sin(d / 10000^{2i/d_{model}}) \qquad (4)$$

$$RPE(i, j, 2d+1) = cos(d / 10000^{2i/d_{model}}) \qquad (5)$$

where (i,j) is the pixel position, and d is the dimension of the position encoding. Additionally, we attempted to normalise the sine positional encoding to accelerate the training. Therefore, linear and sine positional encoding can easily embed row position information into the features.

The linear position encoding assigns linear weights to each row and is directly proposed based on the findings in Figure 1. The sine position encoding was proposed by Aswani et al. (2017); it enables the model to learn the relative position information straightforwardly. Normalised sine position encoding is designed to emphasise the feature information from different rows and ensure smoothness of attention changes.

### 3.3 Loss Function

The network was trained using a multi-task training method that contains four tasks: semantic segmentation, boundary segmentation, and segmentation in the boundary region.

Semantic loss (S-Loss) $L_S$ represents the traditional crossentropy loss, which is formulated as

$$L_S = -\sum_{i \in pixels, c \in classes} (s_{i,c} log(s_{i,c}^{(0)}) + s_{i,c} log(s_{i,c}^{(1)})) \tag{6}$$

where $s_{i,c}$ and $s_{i,c}$ are the prediction and ground truths of the $i_{th}$ pixel for Class c, respectively. This loss is used to evaluate the quality of the classification results of the categories.

Boundary loss (B-Loss) $L_B$ represents the binary cross-entropy loss required to improve the effectiveness of the small-object features, which is formulated as

$$L_B = -\sum_{i \in pixels} (b_i log(b_i) + (1-b_i) log(1-b_i)) \tag{7}$$

where $b_i$ and $b_i$ are the prediction and ground truths of the $i_{th}$ pixel at the boundary. $b_i$ is the result obtained from ground truth $s$ through Canny edge detection. B-Loss evaluates the quality of the boundary prediction results.

The boundary and semantic loss (BAS-Loss) $L_{BAS}$ mixes the final result as follows:

$$L_{BAS} = -\sum_{i \in pixels, c \in classes} \mathbb{I}_{b_i > t} s_{i,c} log(s_{i,c}^{(1)}) \tag{8}$$

where $t$ is a hyperparameter that represents the threshold for the boundary confidence. The BAS-Loss method combines the boundary prediction results with the category classification results, enabling the network to focus on the accuracy of the category in the boundary. The overall loss is formulated as follows:

$$L = L_S + L_{BAS} + L_B \tag{9}$$

### 4. Experimental Setup and Analysis

### 4.1 Experiment Platform

Currently, various studies (Bovcon et al., 2021) have focused on obstacle segmentation methods from USVs; however, limited research has focused on methods from UAVs. The perspective of UAVs outperforms that of USVs, and the results from UAVs are rarely affected by nearby obstacles. Although numerous studies have been dedicated to object detection methods from UAVs, the results from object detection cannot effectively indicate the available navigational area, as indicated by the results from obstacle segmentation.

To collect marine data from USVs and UAVs and to analyse the results between the views of UAVs and USVs, we developed a USV to collect surface data and purchased an advanced UAV to collect aerial data. The overall images of the UAV and unmanned ship are presented in Figure 4. The dimensions of the unmanned ship were 9400 × 4950 × 5352 cm, with a maximum draft of 1 m and a payload capacity of 8.5 tons. The ship was equipped with a high-resolution camera, flexible damping function, satellite communication, and other sensors. The UAV used, DJI Mavic 3, was manufactured by DJI. The DJI Mavic 3 was equipped with a Hasselblad L2D-20c Camera (3840 × 2160), which can collect large amounts of high-resolution aerial information.

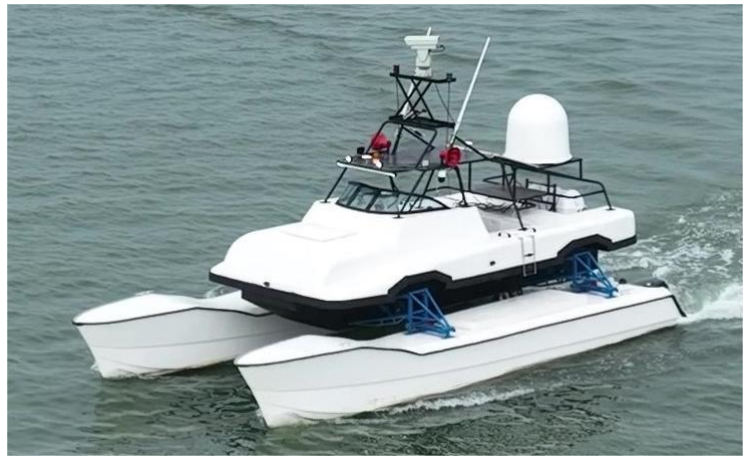

Figure 4. Overall image of the experimental USV

To include more image data, 254 videos were extensively captured in Binhai Port, Yancheng City, Jiangsu Province, and the Yangtze River; the videos spanning a duration of 13,834.9 s containing views of both USVs and UAVs. Each video had a resolution of 3840 × 2160 and a frame rate of 60 fps, which implied that more than 830 k images were available for use and analysis.

**4.2 Our Dataset**

Finally, we selected 1,315 images for annotation and merged them into our dataset. These data were collected based on several dimensions to ensure completeness and scientific rigour of data collection. Data from the USV perspective and at different flying heights (25, 50, and 100 m) were extensively collected. The pitch angles of the collected data were categorised as 0, 10, and 20˘r. Approximately half of our data focused on the semantic segmentation of the sea and sky, while the other half focused on the semantic segmentation of the sea and sky. Additionally, the data were collected based on the time (morning, afternoon, and nightfall) and offshore distance (close-, medium-, and long-range). To improve the completeness and scientific rigour of the dataset, we supplemented the close-up information of the ships and departure records from the USV perspective.

To completely utilise all data, polygons were drawn to indicate the sky, sea, land, obstacles, towers, and ships. We labelled the tower tag because ports often have numerous engineering ships, and their features differ from those of regular ships. Therefore, we labelled the engineering ship platform as a ship and the engineering equipment as a tower.

To annotate the data accurately and reduce errors, we followed the annotation method proposed by MaSTr1325 (Bovcon et al., 2019). To accelerate the annotation process, the brush was adjusted to multiple polygons. Moreover, to improve the annotation speed, Eiseg (Liu et al., 2021; Contributors, 2019), an efficient and intelligent interactive segmentation annotation software, was used during the annotation process. In the practical experiments, the annotation process and quality control required an average of 5 min per image. Through manual annotation, 1,315 images were annotated, including 1,598 sea polygons, 1,327 sky polygons, 1,231 land polygons, 4,253 obstacle polygons, 1,856 tower polygons, and 7,965 ship polygons (as shown in Figure 5). A significant difference between the data collected by the UAV and USV was obtained. The higher the altitude, the wider the field of view; however, the separation between the sea and sky blurred.

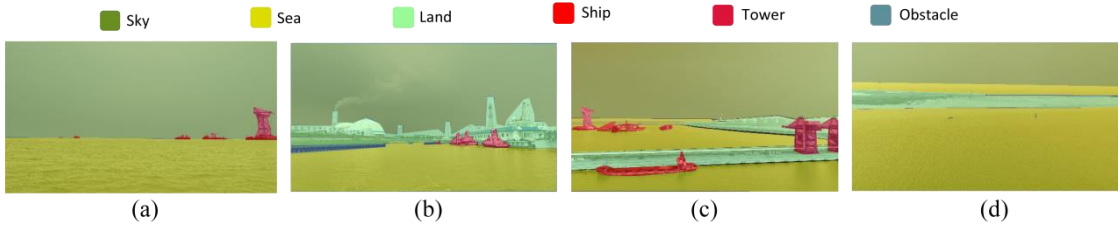

(a) (b) (c) (d)

Figure 5. Annotated examples of our dataset. (a) and (b) are captured by the USVs. (c) and (d) are captured by the UAVs.

**4.3 Evaluation Criteria and Implementation Details**

Following the evaluation criteria of MaSTr1315 (Bovcon et al., 2019), the accuracy of the semantic segmentation was evaluated using five metrics. The intersection over union metric (IOU) can be formulated as follows:

$$IOU_c = \frac{pred_c^c}{gt_c + \sum_{k \in classes} pred_c^k - pred_c^c} \quad (10)$$

$$mIOU = \frac{1}{num\_classes} \sum_{c \in classes} IOU_c \quad (11)$$

Where $gt_c$ is the pixel number of the $c_{th}$ class, and $pred_i^j$ represents the number of pixels of the $j_{th}$ class predicted to pertain to the $i_{th}$ class. The IOU metric calculates the ratio of the intersection area between the predicted and ground truth regions to the union area of both regions. This metric was used to measure the degree of overlap between the predicted and true regions. Bovcon et al. (2019) proposed that the IOU metric is the appropriate metric to evaluate the prediction accuracy between the sea and sky.

Moreover, the accuracy metric (ACC) calculates the ratio of correctly classified samples to the total number of samples. This metric is used to measure the prediction accuracy; it can evaluate the prediction accuracy of small obstacles. In addition, Bovcon et al. (2018) found that the F-measure can effectively assess a model's predictive performance for obstacles.

In the experiments, we used PIDNet-s as the baseline. In particular, the STEM network uses four residual modules with a kernel size of three and a stride of two. The RPEM in the detail branch contains two residual modules with a kernel size of three and a padding size of one to ensure that the input and output shapes are the same. The poly strategy was used as the learning rate updating strategy during training. We used random re-sizing, random cropping, and random horizontal flipping for data augmentation and set the training batch size to six. The cropped image size was 536 × 960. The network was trained for 120k iterations, and the initial learning rate was 0.01. The network was implemented in mm segmentation (Contributors, 2020) and ran on a platform with a Core i7-13700K CPU, a single NVIDIA RTX 4070, Pytorch 1.10, CUDA 11.7, and Linux-Conda environment. The final training period was 14 h. During the test period, we used the tools in mmengine (Contributors, 2022) to calculate the floating point operations per second (FLOPs) and the number of parameters (Params) of the network.

**4.4 Comparative Experiments**

Comparative experiments were conducted to evaluate the performance of the baseline network PIDNet and the network with RPEM on MaSTr1325 and our dataset. According to the position encoding differences, RPEM-linear, RPEM-sine, and RPEM-sin-norm represent the RPEM with linear position encoding, sine position encoding, and normalised sine position encoding, respectively. Moreover, we visualised the baseline and our network results to demonstrate the superiority of our network. From these results, the advantages of the baseline network with RPEM compared with the baseline network without RPEM can be evaluated.

**4.4.1 Results on MaSTr1325**

To demonstrate the efficiency of the RPEM, we applied our three-branch semantic segmentation network to the RPEM on MaSTr1325 (Bovcon et al., 2019), which is highly similar to our dataset. The baseline network, PIDNet, was compared with our network. As summarised in Table 1, the network with RPEM outperformed its baseline network in predicting the sky and sea. The prediction accuracy of the sky and sea is more significant than that of other classes because most parts of maritime images correspond to the sea and sky region; thus, distinguishing between the sea and the sky affects the USV safety navigation.

From the results presented in Table 1, the network with RPEM-linear surpassed its baseline model PIDNet-s and the improved baseline model PIDNet-l in terms of the IOU and ACC metrics. Compared to the baseline model PIDNet-s, the network with RPEM-linear outperformed the mIOU, mACC, and aACC metrics by 6.88%, 3.94%, and 4.94%, respectively. Even for the improved baseline model PIDNet-l, the network with RPEM-linear outperformed the mIOU, mACC, and aACC metrics by 5.84%, 3.73%, and 4.65%, respectively. Particularly, for the sea and sky categories, all models with RPEM-sin, RPEM-sin-norm, and RPEM-linear showed an improvement of at least 3.69% and up to 11.4% IOU compared with the baseline.

### 4.4.2 Results on our dataset

The results, summarised in Table 2, are similar to those for the public datasets. In our dataset, the network with the RPEM module consistently outperformed the baseline model. The best-performing model showed improvements across mIOU, mACC, and Fscore metrics. Compared with the baseline model PIDNet-s, the network with RPEMsin outperformed the mIOU, mACC, and aACC metrics by 1.04%, 0.67%, and 1.22%, respectively. Even for the improved baseline model PIDNet-l, the network with RPEM-linear outperformed the mIOU, mACC, and aACC metrics by 0.31%, 0.38%, and 0.12%, respectively. Moreover, even for the obstacle category, which is less associated with row indices, the network with RPEM at least maintained the performance of the baseline network in the F-score metric.

Notably, the performance of the model with RPEM regarding the sky and sea categories significantly surpassed that of the baseline model and even exceeded the performance of the enhanced version of the baseline model. For the sea and sky categories, all models with RPEM-sin, RPEM-sin-norm, and RPEM-linear showed an improvement of at least 1.25% and up to 3.83% in the IOU compared to the baseline. Our model performed relatively poorly compared to the baseline model for the obstacle and ship categories. This is because no strong correlation was present between the obstacle and ship categories and the row coordinates. However, in practical applications, obstacles are only worth considering when they are close to a ship, and in such instances, they are often detectable. However, the accuracy of distinguishing between the sea and the sky affects the calculation of the attitude of the ship and other crucial information. Therefore, we should focus on the prediction accuracy of sea and sky categories.

**4.4.3 Analysis**

To prove the effectiveness of RPEM, we analysed the results of PIDNet-s, PIDNet-l, and PIDNet-s with RPEM-sin in Figure 6. The results indicate that the network with RPEM effectively alleviated the misclassification problem between the sea and the sky. In the PIDNet-s and PIDNet-l results, a large portion of the pixels in the sky were misclassified as sea. However, PIDNet-s with RPEM-sin effectively alleviated this problem and showed improved capability for discriminating between the sea and the sky. Moreover, the boundary generated by the network with RPEM between the sea and the sky was smoother than that of the baseline model. The results in rows 2–4 show fluctuations in the predicted boundary between the sea and the sky in PIDNet-s and PIDNet-l, whereas PIDNet-s with RPEM-sin exhibited a comparatively smoother transition. This indicates that our model is robust in predicting the sea and sky categories. To illustrate the advantage of the model in distinguishing between the sea and sky, we visualised the confidence distribution of the sea for the selected images, as shown in Figure 7. The data from the first row indicate that PIDNet-s with RPEM-sin can significantly improve the prediction confidence of marine regions, and the data from the second row indicate that PIDNet-s with RPEM-sin can alleviate misclassifying the sky as the sea. The results indicate that our model exhibits higher discrimination between the practical sea and sky regions regarding the prediction scores for the sea class.

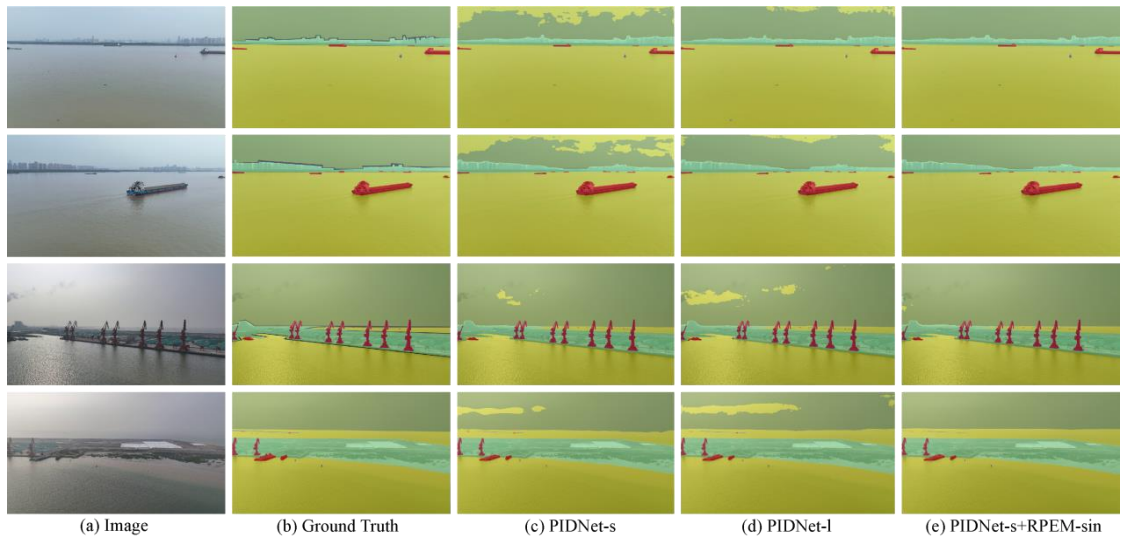

(a) Image    (b) Ground Truth    (c) PIDNet-s    (d) PIDNet-l    (e) PIDNet-s+RPEM-sin

Figure 6. Segmentation visualisation of PIDNet-s, PIDNet-l, and PIDNet-s with RPEM-sin. The maps in each row from left to right are the origin image, ground truth, and results of PIDNet-s, PIDNet-l and PIDNet-s with RPEM-sin.

Although the RPEM effectively enhances the ability of networks to distinguish between the sea and the sky, there are still some limitations. For example, the prediction accuracy for fine and small objects is poor, as shown in Figure 8(a). Because an object occupies fewer pixels, the boundary prediction for small objects can include the surrounding non-edge areas in the three-branch network. Moreover, in a maritime port environment, most towers are crane towers without antennas, leading to imbalanced samples. Figure 8(b) shows the confusion between the sea and sky. This is because the data in maritime port environments frequently exhibit a sky–sea–land–sea structure. Even with a specialised network structure, this may lead to confusion between the sea and the sky. In Figure 8(c), small obstacles at a distance near the waterline can significantly affect sea–sky discrimination. When two objects are significantly close, the information at the boundary area is relatively high, which may be mistaken as noise, thereby blurring the boundaries between objects. Figure 8(d) shows the confusion between the sky and the tower. This is because the clouds in the sky sometimes affect the network owing to their unique shapes.

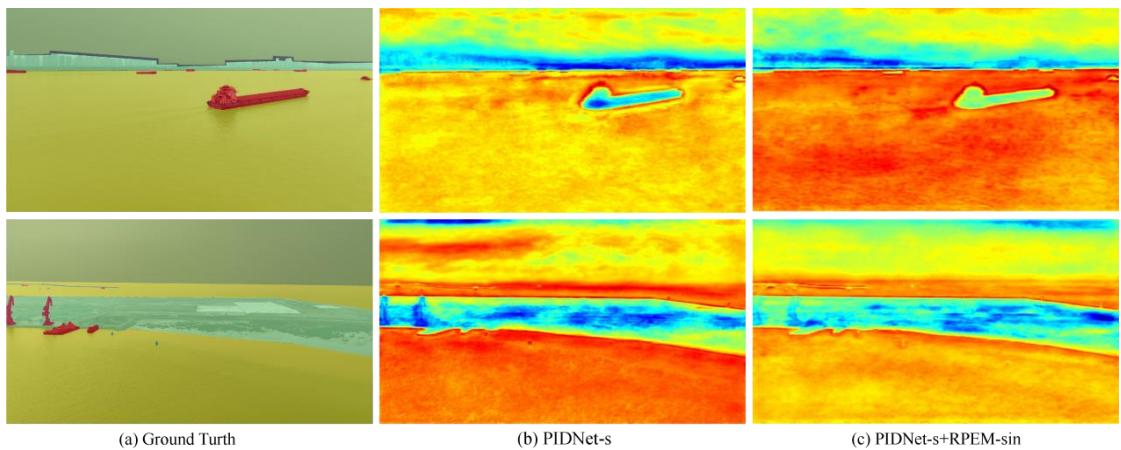

Figure 7: Heatmap of the sea. The more intense the red colour at a particular pixel, the higher the confidence score, indicating the prediction of that pixel as pertaining to the sea class.

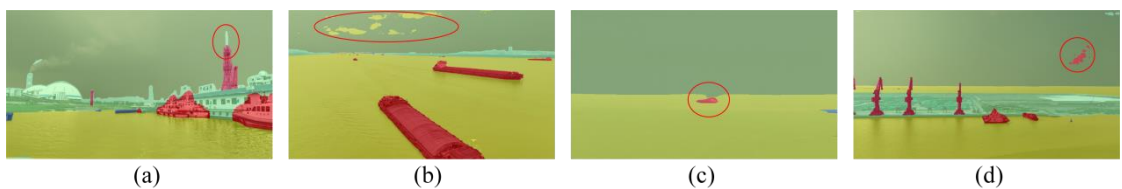

Figure 8: Error results of PIDNet-s with RPEM-sin on our dataset.

Table 1. Results on MaSTr1325. **Bold** text indicates the best performance.

| Class | IOU obstacle | sea | sky | mIOU All | ACC obstacle | sea | sky | mACC All | aACC All | F-score obstacle | sea | sky | mF-score All |
|---|---|---|---|---|---|---|---|---|---|---|---|---|---|
| **PIDNet-l** | 92.44 | 91.47 | 94.33 | 92.75 | **98.59** | 91.65 | **99.81** | 96.68 | 96.4 | 96.07 | 95.55 | 97.08 | 96.23 |
| **PIDNet-s** | **93.05** | 85.83 | 90.27 | 89.72 | 97.99 | 86.22 | 99.64 | 94.62 | 94.06 | 96.4 | 92.37 | 94.88 | 94.55 |
| **+RPEM-noise** | 93.29 | 87.98 | 91.58 | 90.95 | 97.96 | 88.18 | 99.76 | 95.3 | 94.92 | **96.53** | 93.6 | 95.6 | 95.25 |

| | | | | | | | | | | | | |
|---|---|---|---|---|---|---|---|---|---|---|---|---|
| +RPEM-sin | 92.64 | 91.75 | 93.96 | 92.78 | 96.66 | 91.98 | 99.79 | 96.14 | 96.36 | 96.18 | 95.7 | 96.89 | 96.25 |
| +RPEM-sin-norm | 92.87 | 94.19 | 96.11 | 94.39 | 98.19 | 94.41 | 96.11 | 97.46 | 97.48 | 96.31 | 97.01 | 98.02 | 97.11 |
| +RPEM-linear | 91.83 | **97.23** | **98.59** | **95.89** | 97.6 | **97.94** | 99.51 | **98.35** | **98.71** | 95.74 | **98.6** | **99.29** | **97.88** |

Table 2. Results on our dataset. **Bold** text indicates the best performance.

| | IOU | | | | | | mIOU | ACC | | | | | | mACC | aACC | mF-score |
|---|---|---|---|---|---|---|---|---|---|---|---|---|---|---|---|---|
| Class | sea | sky | land | obstacle | tower | ship | All | sea | sky | land | obstacle | tower | ship | All | All | All |
| FCN | 90.46 | **93.69** | 67.57 | 42.28 | 58.15 | 60.57 | 68.79 | 91.4 | 98.76 | 91.59 | **52.39** | 80.35 | 83.13 | 82.93 | **93.38** | 80.13 |
| +RPEM-linear | **91.06** | 86.31 | **77.97** | 42.87 | 56.64 | 69.51 | 70.72 | **92.13** | 96.7 | 91.83 | 48.95 | 81.86 | 84.54 | 82.67 | 93.20 | 81.66 |
| +RPEM-sin-norm | 90.38 | 85.58 | 82.22 | 39.19 | 60.75 | 71.81 | 71.65 | 91.36 | 97.74 | 92.51 | 48.12 | 82.8 | **85.28** | 82.97 | 93.17 | 82.15 |
| +RPEM-sin | 89.42 | 87.33 | 77.8 | **44.58** | **66.78** | **72.63** | **73.09** | 90.25 | **98.79** | **93.76** | 52.35 | **84.08** | 83.73 | **83.83** | 92.97 | **83.51** |
| PIDNet-l | 97.59 | 96.04 | 94.25 | **68.89** | 85.91 | **89.48** | 88.69 | 99.3 | 97.14 | 96.83 | **79.18** | 91.74 | 94.35 | 93.09 | 98.21 | 93.71 |
| PIDNet-s | 96.39 | 93.82 | 94.27 | 68.07 | **86.83** | 88.37 | 87.96 | 99.28 | 94.77 | **96.98** | 78.26 | 92.76 | **94.71** | 92.8 | 97.49 | 93.3 |
| +RPEM-linear | 97.64 | 96.6 | 93.07 | 58.37 | 80.88 | 85.54 | 85.35 | 99.09 | 97.79 | 96.77 | 67.56 | 92.81 | 89.63 | 90.61 | 98.18 | 91.47 |
| +RPEM-sin-norm | 98.08 | 97.16 | 94 | 68.53 | **86.83** | 88.56 | 88.86 | 99.34 | 98.22 | 96.5 | 76.45 | 92.23 | 93.52 | 92.71 | 98.51 | 93.78 |
| +RPEM-sin | **98.42** | **97.65** | **94.43** | 67.4 | 87.25 | 88.84 | **89** | **99.44** | 98.29 | 97.5 | 78.38 | **92.97** | 94.21 | **93.47** | **98.71** | **93.83** |

### 4.5. Ablation Experiments

Ablation experiments were conducted to evaluate the RPEM performance. We deployed the RPEM on the traditional semantic segmentation network FCN (Long et al., 2015) to demonstrate its effectiveness in networks with different structures. We also compared the effects of RPEM with different positional encodings on different datasets.

#### 4.5.1 Effectiveness of RPEM

To prove the efficiency of position encoding, the network with the RPEM was compared with the network embedded with noise encoding, which only replaces the row position encoding with Gaussian noise in the RPEM and retains the original computational pathways of the RPEM. The results are listed in Table 2 and indicate that the FCN with the RPEM outperformed baseline model PIDNet-s in classifying the sea and sky by 4.3%, 0.37%, and 3.38% in the mIOU, mACC, and mF-score metrics, respectively.

**4.5.2 Effectiveness of designed position encoding**

Owing to the increased computational workload, the network with noise performed better than the baseline; however, its performance in the sea and sky was significantly worse than that of the network with RPEM. Moreover, by further increasing the number of parameters and forming an improved baseline model (PIDNet-l), the performance can be further improved. This result indicates that increasing the computational and parameter complexity can improve the model's performance; however, RPEM with designed position encoding can achieve better performance with fewer additional computational and parameter requirements.

**4.5.3 Effectiveness of different positional encoding**

Based on the results of our dataset and MaSTr1325 listed in Tables 1 and 2, different encodings exhibited variations in performance on different datasets. On our dataset, sine position encoding performed better. However, for the MaSTr1325 dataset, linear position encoding performed better. Furthermore, normalised sine position encoding consistently performed as the second-best performer in our dataset and MaSTr1325. These findings indicate that the choice of position encoding affects the performance of the RPEM according to the application environment; therefore, the selection of position encoding should be based on the specific application environment.

**4.6. Inference Speed Experiments**

To compare the speed difference between the baseline and the network with RPEM, we calculated the FLOPs and the Params of the baseline model, improved the baseline model, and the model with RPEM. From the results presented in Table 3, the Params of the model with RPEM increased by only 2.7% compared to the baseline model. However, it performed better than the improved baseline model, whose Params increased by 383.4%. Additionally, the FLOPs of the model with RPEM increased by only 4.4%, whereas the FLOPs of the improved baseline model increased by 480%. These results indicate that the model with the RPEM achieved better performance using less storage and computational resources.

Table 3. Comparison of speed and accuracy on our dataset

|  | Flops | Params | Size |
|---|---|---|---|
| **PIDNet-s** | 11.719G | 7.718M | $536 \times 960$ |
| **+RPEM** | 12.239G | 7.739M | $536 \times 960$ |
| **PIDNet-l** | 67.97G | 37.308M | $536 \times 960$ |

## 5. Conclusion

This study presented a novel RPEM to consider the effect of row coordinates on features and constructs a three-branch semantic segmentation network with RPEM for maritime port semantic segmentation tasks. To the best of our knowledge, this is the first study on a maritime port environment semantic segmentation task from both USV and UAV perspectives. Our model effectively mitigates the problem of distinguishing sea from sky and predicts smoother and more robust boundaries. However, the performance of the network with RPEM depends on the selection of the position encoding in different application environments. Moreover, the network still exhibits limitations in predicting small obstacles, such as buoys at long distances. In future work, we will explore more effective encoding techniques to address the dependency of encoding on specific application scenarios and use a self-supervised method to alleviate the issue of limited samples.

**Acknowledgements**

This study was supported by the National Training Program of Innovation and Entrepreneurship for Undergraduates of China [Grant number S202310698085].